\pdfoutput=1
\documentclass[11pt]{article}

\usepackage{EMNLP2023}

\usepackage{tabularx}
\usepackage{tabulary}
\usepackage{multirow}
\usepackage{tabularray}
\usepackage{inconsolata}
\usepackage{graphicx}
\usepackage{booktabs}
\usepackage{enumitem}
\usepackage{lipsum}

\usepackage{times}
\usepackage{latexsym}

\usepackage{graphicx}
\usepackage{siunitx,booktabs,caption}

\title{The Balancing Act: Unmasking and Alleviating ASR Biases in Portuguese}

\author{Ajinkya Kulkarni  \\
  MBZUAI, UAE \\
\texttt{ajinkya.kulkarni@mbzuai.ac.ae} \\
  \newline \\
\textbf{Rameez Qureshi}\\
  ADAPT Centre, Trinity College Dublin  \\
\texttt{rameez.qureshi@adaptcentre.ie} \\
    \\\And
Anna Tokareva  \\
  University of Lorraine\\
  \texttt{anna.tokareva3@etu.univ-lorraine.fr} \\
  \newline \\
\textbf{Miguel Couceiro}\\
  University of Lorraine, CNRS, LORIA\\ \texttt{miguel.couceiro@loria.fr}}

\begin{document}
\maketitle

\begin{abstract}

In the field of spoken language understanding, systems like Whisper and Multilingual Massive Speech (MMS) have shown state-of-the-art performances. This study is dedicated to a comprehensive exploration of the Whisper and MMS systems, with a focus on assessing biases in automatic speech recognition (ASR) inherent to casual conversation speech specific to the Portuguese language. Our investigation encompasses various categories, including gender, age, skin tone color, and geo-location. Alongside traditional ASR evaluation metrics such as Word Error Rate (WER), we have incorporated p-value statistical significance for gender bias analysis. Furthermore, we extensively examine the impact of data distribution and empirically show that oversampling techniques alleviate such stereotypical biases. This research represents a pioneering effort in quantifying biases in the Portuguese language context through the application of MMS and Whisper, contributing to a better understanding of ASR systems' performance in multilingual settings.

\end{abstract}
%Please keep the "track changes' on until the paper is finalized 

\section{Introduction}

%a. Current SOTA in Multilingual ASR, and Spanish, Portuguese
%b. Current aspects of Fairness in ASR and effect on socio-economics and cultural etc etc
%c. Why it is important to study biases in terms of causal conversations and their impact 
% Previous work on bias in ASR, and multilinguality
%d. Main contributions of this work:
%    A. First study wrt to skin-tone, Age, Gender and Locations
%    B. Discovering biases in Large-acoustic models
%    C. Detailed analysis on conversational speech

Conversational Artificial Intelligence (AI) has become increasingly integrated into everyday applications over the past few years. The history of previous broad technologies shows that despite temporary challenges, restructuring the economy around innovative technologies offers significant long-term benefits \cite{TheLongandShortofTheDigitalRevolution}. 
%Hence, it is wise to offer 
This asks for fair AI solutions that can connect people from different backgrounds, and that enables
%enabling 
universal access to technology. In the context of human-machine interactions through spoken language, Automatic Speech Recognition (ASR) facilitates smooth information exchange within various conversational AI applications, including machine translation, sentiment analysis, and question-answering systems \cite{Bangalore2005IntroductionTT}.

The significance of spoken language in our daily lives emphasizes the need for ASR systems to accommodate the various forms of human communication. It is thus vital that ASR systems can adeptly manage this diversity, as it is crucial for enabling smooth and inclusive communication across a wide range of situations and people, and extending the use of ASRs in domains such as emergency services, home automation, and navigation systems. To accommodate fairness and transparency requirements 
it is %pertinent to initially 
 paramount to examine the prevailing biases within various subgroups towards fair ASR systems.

Over the past few years, there has been a growing research community examining biases in automatic speech recognition (ASR) systems \cite{rac1,rac2,rac3,rac4,LimaFFA19,BlodgettBDW20}. This research has primarily focused on assessing the impact of disparities related to gender, age, accent, dialect, and racial meta-attributes. (It is worth mentioning that most of these features are considered sensitive according to legal protection against discrimination, {\it e.g.,}  in the U.S.\footnote{\url{https://www.whitehouse.gov/ostp/ai-bill-of-rights/##applying}} and in Europe\footnote{\url{https://ec.europa.eu/newsroom/dae/document.cfm?doc_id=60419}}.) However, the majority of these studies have been carried out on monolingual ASR systems for the English language, with only a limited number of studies addressing bias detection in non-English languages.

In the study conducted in \cite{du_bias1,du_bias2}, researchers examined the (Hidden Markov Model) HMM- Deep Neural Network (DNN) ASR system to assess biases related to gender, age, and accents in the context of the Dutch language. They then proposed the use of data augmentation and vocal tract length normalization techniques to alleviate these biases in Dutch ASR systems \cite{du_bias3}. Another study centered on French broadcasting speech, aimed to uncover gender biases and revealed that the under-representation of specific gender categories could result in bias in HMM-DNN ASR performance, regardless of gender identity (male, female, or other) \cite{fr_bias1,fr_bias2}. Furthermore, it emphasized the importance of a systematic examination of demographic imbalances present in datasets. 

For Arabic ASR system, which were developed using Carnegie Mellon University Sphinx 3 tools\footnote{{\url{https://www.cs.cmu.edu/~archan/sphinxInfo.html}}}, an investigation was conducted to understand the impact of gender, age, and regional factors on performance \cite{ar_bias1}. While these studies laid the foundation for quantifying biases, there remains a scarcity of research on ASR systems trained with large amounts of multilingual data, even though they consistently achieve state-of-the-art performance levels.

The emergence of computational resources enabled the acceleration of the development of large pre-trained acoustic models, resulting in unified frameworks with multilingual capabilities. These frameworks are often built upon transformer networks and prominently use the Wav2vec 2.0 \cite{w2vec} framework. As a consequence, there has been a significant push to create multilingual ASR systems\cite{asr2k,whisper,ggleusm,mms}, extending their applicability to more than 100 languages, including those with limited linguistic resources. Meta AI's MMS system \cite{mms} conducted an evaluation that included the assessment of gender and language biases using the FLEURS dataset \cite{fleurs}. However, there is still a need for a deeper understanding of the comparative differences among various multilingual ASR systems when it comes to quantifying potential biases.

To explore the biases present in multilingual ASR systems trained on extensive speech data, we investigated variants of OpenAI's Whisper ASR system \cite{whisper} and Meta AI's MMS ASR system \cite{mms}, both of which have achieved state-of-the-art performance levels. In addition, we selected the Casual Conversation Dataset version 2 (CCD V2) to quantify biases and assess the fairness of these system performances in the context of the Portuguese language \cite{ccdv2}. Our study takes into account a diverse spectrum of categories, including age groups, gender, geographical locations, and skin tones. The consistency in textual content across all CCD V2 recordings establishes a robust basis for the efficient evaluation of system performance across a broad array of categories. Only a limited number of studies have delved into the influence of state-of-the-art multilingual ASR systems on domain-specific ASR tasks. For example, these studies have explored code-switching between languages using systems like Whisper and MMS \cite{stud1}, or they have examined the effects of ASR errors on discourse models among groups of students in noisy, real-world classroom settings between Whisper and Google ASR system \cite{stud2}. %may remove last 2 lines if needed

More often, an imbalanced distribution of evaluation data across various sub-categories can result in an inadequate analysis of the evaluation process itself. Therefore, we %have introduced this study along with 
explore two resampling methods, namely, {\it naïve} and {\it Synthetic Minority Oversampling Technique} (SMOTE)\cite{Smote}, to ensure a balanced data distribution across each subgroup when quantifying the biases. In the assessment of ASR systems, our primary choice of metrics includes Word Error Rate (WER) and Character Error Rate (CER)\footnote{In this paper, we only include the WER results. The CER results are provided in \url{https://biasinai.github.io/asrbias/}.}. Interestingly, we observe that oversampling techniques can alleviate performance disparities between certain subgroups. % complemented by the examination of disparity ratios. 
%Additionally, to determine whether distinctions among various subgroups hold statistical significance, we conducted p-value tests, using a significance level set at a threshold value of 0.05.

The structure of the paper is as follows. In Section 2, we provide an overview of the Casual Conversation dataset, which is utilized to quantify biases in multilingual ASR systems in the Portuguese language. We described the specifics of the MMS ASR system and the variants of Whisper ASR systems along with the evaluation protocol in Section 3.  We outline results along with an analysis on various categories to to quantify biases in Section 4, along with the corresponding evaluation methodologies. Section 5 details the discussion, and we draw our conclusions in Section 6 along with potential directions for future work. 

The \textbf{main contributions} of this paper are as follows: 
\begin{enumerate}
\item It presents the first study on analyzing disparities within multilingual ASR systems focused on the Portuguese language.
\item It emphasizes the critical significance of data distribution among sub-categories by employing oversampling techniques.
\item It illustrates the comparative distinctions between Whisper ASR and MMS ASR, and examines the impact of model parameters on the development of an efficient system design.
\item In addition to gender and age groups, it investigates skin tone and geo-location as criteria to measure inter-racial biases.
%disparity impacting the 
\end{enumerate}

\section{Dataset Description}

The CCD V2 dataset is open-source and can be accessed through the Meta AI website\footnote{https://ai.meta.com/datasets/casual-conversations-v2-dataset/}. It %was carefully curated and 
represents the speech of 5,567 unique speakers from various regions, including India, the United States of America, Indonesia, Vietnam, Brazil, Mexico, and the Philippines. This compilation results in five audio samples per individual, yielding a total of 26,467 video recordings. The dataset encompasses seven self-labeled attributes, including details about the speaker's age, gender, native and secondary languages or dialects, disabilities, physical characteristics, and adornments, as well as geographic location. Additionally, it features four other characteristics: two skin tone scales (Monk Skin Tone \cite{monkskin} and Fitzpatrick Skin Type \cite{fitzpatric1,fitzpatric2}), voice timbre, the speaker's activity, categorized as gesture, action, or appearance, and details about the recording setup, which covers video quality, background environment, and video configuration. For Monk skin tone scale-10 only one sample was available for Portuguese language. Therefore, in order to avoid skewed comparison between skin-tone scales using Monk skin tone, we only conducted a study using Fitzpatrick skin type. 

The CCD V2 comprises 354 hours of recordings where speakers responded to specific questions in a non-scripted manner and 319 hours of recordings in which individuals read passages from F. Dostoyevsky's ``The Idiot'', translated into various languages. Throughout this paper, we utilized scripted recordings for the Portuguese language. As each scripted recording had the same textual content and phonetic variations, it enables the examination of meta-attributes leading to performance differences. 
%In the course of this paper, we made use of scripted recordings in the Portuguese language. 
For more comprehensive details of CCD V2 and the dataset design process, please refer to the works published in \cite{ccdv2} and \cite{ccdv2other}.

%The data is distributed in several zip archives, each approximately 30-60 GB in size, along with corresponding transcriptions for scripted recordings and annotations in JSON format. 

In the context of assessing the fairness of ASR systems, we focused primarily on a subset of scripted recordings, with a strong emphasis on the Portuguese language. In this study, we concentrated on four annotated labels: gender, age, Fitzpatrick scale, and geographic location. To simplify our analysis, we categorized speakers into seven age groups: 18-24, 25-30, 31-36, 37-42, 43-50, 51-60, and 61+. 
After the initial analysis of the evaluation sample distribution for each sub-category, we observed imbalanced distributions among various subgroups. We thus explored resampling strategies to ensure that biases are not introduced into the computed results due to imbalanced distributions across subgroups.

\section{Empirical study}

%[Put a short sentence to introduce the contents of the section]

In this empirical study, we initiate our investigation by conducting a thorough analysis of the influence of various sampling techniques on performance disparities within multilingual ASR systems for Portuguese. Additionally, in Section 3.1, we first present the ASR systems employed in this research. Subsequently, we outline the evaluation protocol and data preparation in Section 3.2. %Lastly, we describe the analysis of disparities related to gender, skin tone, age groups, and geographic location.

\subsection{ASR Systems} 

This study centers around the utilization of state-of-the-art, open-source multilingual ASR systems, specifically Whisper and the Multilingual Massive Speech Systems. Both of these systems have demonstrated their efficacy in a range of speech-processing tasks, including audio classification, speech translation, and text-to-speech synthesis. They have been trained on extensively large-scale multilingual datasets using self-supervised and multi-task learning techniques, enabling support for over 100 languages.

\subsubsection{Whisper}

Whisper \cite{whisper} is a robust speech recognition model presented by OpenAI\footnote{https://openai.com/research/whisper} in 2022. Whisper is trained using a multitask learning on 680,000 hours of labeled multilingual recordings collected from the Internet, along with the corresponding transcriptions filtered from machine-generated ones. In total 96 languages are covered by approximately 117,000 hours of audio data, making Whisper a powerful tool for multilingual speech recognition. 

Whisper incorporates the Transformer encoder-decoder architecture \cite{transformer} with the implementation of multitask learning techniques allowing language identification, multilingual speech transcription, along with word-level timestamps. The input audio is split into thirty-second chunks, which makes the transcription of long recordings more effective. In the Whisper framework, the encoder processes log Mel spectrogram inputs, generating relevant features for the decoder. The decoder, in turn, consumes these encoder features, positional embeddings, and a sequence of prompt tokens. Subsequently, it produces the transcribed text corresponding to the input speech.

Whisper has different variants based on model parameter sizes such as Tiny (39 Million), Base (74 Million), Small (244 Million), Medium (769 Million), Large (1550 Million), and Large-v2 (1550 Million). Whisper models are primarily divided into two categories based on languages and tasks: English-only models and multilingual models. In this paper, we incorporated Medium, Large, and Large-v2 variants of Whisper. 

\subsubsection{Massively Multilingual Speech system}

In 2023, Meta AI released the Massively Multilingual Speech (MMS) project, as documented in \cite{mms}, expanding its language support to encompass over 1000 languages for various speech processing applications. The primary components of the MMS system include a novel dataset derived from publicly accessible religious texts and the adept use of cross-lingual self-supervised learning. 
The MMS project encompasses various tasks, such as speech recognition, language identification, and speech synthesis. MMS is built upon the Wav2Vec 2.0 \cite{w2vec} architecture and has undergone training through a combination of cross-lingual self-supervised learning and supervised pre-training for ASR. It incorporates language adapters that can be dynamically loaded and interchange during inference, featuring multiple Transformer blocks, each augmented with a language-specific adapter.

The authors compiled two datasets using texts from the New Testament and the Bible, along with recordings of readings of these religious texts available on the Internet. The labeled dataset (MMS-lab) comprises 1,306 audio recordings of New Testament readings in 1,130 languages, resulting in 49,000 hours of data and approximately 32 hours of data per language. The audio underwent several alignment stages, including training several alignment models and a final filtering of noisy or paraphrased data. The unlabeled dataset (MMS-unlab) contains 9,345 hours of audio and includes recordings collected from the Global Recordings Network, organized into 3,860 languages. The MMS system is available in two variants based on model parameters, with 317 million and 965 million parameters. For this study, we utilized the MMS system with 965 Million model parameters.

\subsection{Preprocessing and evaluation processes}

In this subsection, we will first outline the pre-processing steps employed to prepare the evaluation dataset using CCD V2 for Portuguese. We will explain the sampling methods for analyzing biases within sub-categories and subsequently discuss the evaluation measures used to assess disparities among these sub-categories. 

\subsubsection{Handling imbalance}

Imbalanced evaluation data can have a detrimental effect on the results, making it challenging to discern meaningful distinctions between the groups being compared. From Table \ref{tab:transposed_table_complete}, we observe that initially collected samples for Portuguese have unbalanced distributions across several categories, which may impact the assessment of ASR systems towards measuring disparities towards underrepresented classes. Therefore, we opted for data balancing approaches, specifically focused on oversampling, and subsequently compared the results. 

It is also worth mentioning that after preliminary analysis of ASR systems results, we observed that the Portuguese subset of the CCD V2 dataset contains audio recordings named "Portuguese scripted" but representing the speech of people speaking on various topics but not reading the passage from Dostoevsky's novel. This might have been a mistake during the compilation of the CCD V2 dataset. These samples were deleted from our evaluation data since the WER for the corresponding transcriptions was exceptionally high and negatively affected the overall performance. 

\begin{table*}[ht]
\centering
\fontsize{9.5pt}{9.5pt}\selectfont
\resizebox{\textwidth}{!}{% Resize table to fit within \textwidth horizontally
\tiny % Use tiny font size for the table content
\begin{tabular}{|l|l|l|l|l|l|l|l|l|l|l|l|l|l|l|l|l|l|l|l|l|l|l|l|l|l|l|l|}
\hline
\multicolumn{2}{|c|}{} & \multicolumn{2}{c|}{Gender} & \multicolumn{6}{c|}{Fitzpatrick scale} & \multicolumn{7}{c|}{Age Groups} & \multicolumn{11}{c|}{Geo-location} \\ \cline{1-28}
\multicolumn{2}{|c|}{} & Male & Female & T.1 & T.2 & T.3 & T.4 & T.5 & T.6 & 18-24 & 25-30 & 31-36 & 37-42 & 43-50 & 51-60 & 61+ & MA & MT & RN & GO & PI & RS & RJ & SP & PE & PR & MG \\ \hline
\multicolumn{2}{|l|}{Initial} & 240 & 500 & 11 & 192 & 289 & 159 & 72 & 17 & 83 & 201 & 164 & 137 & 103 & 44 & 8 & 9 & 27 & 25 & 11 & 7 & 28 & 130 & 379 & 38 & 55 & 31 \\ \hline
\multicolumn{2}{|l|}{Naïve} & 1019 & 1009 & 25 & 681 & 743 & 345 & 164 & 70 & 293 & 282 & 297 & 293 & 283 & 274 & 285 & 18 & 47 & 39 & 24 & 34 & 70 & 409 & 1110 & 59 & 119 & 99 \\ \hline
\multicolumn{2}{|l|}{SMOTE} & 4443 & 4443 & 1925 & 1893 & 2630 & 1132 & 322 & 984 & 1014 & 2918 & 1703 & 1236 & 978 & 458 & 579 & 892 & 854 & 845 & 841 & 830 & 805 & 796 & 787 & 769 & 744 & 723 \\ \hline
\end{tabular}
}
\caption{Statistical representation of samples for demographic categories across Initial, Naïve, and SMOTE datasets. The abbreviations for `Geo-location' are as follows: RN - Rio Grande do Norte, SP - Sao Paulo, RS - Rio Grande do Sul, GO -Goias, MT - Mato Grosso, PR - Parana, RJ - Rio de Janeiro, MG - Minas Gerais, PI - Piaui, PE - Pernambuco, MA - Maranhao. The abbreviations for `Fitzpatrick scale'  are as follows: T.1 - type i, T.1 - type ii, T.1 - type iii, T.1 - type iv, T.1 - type v, T.1 - type vi.}
\label{tab:transposed_table_complete}
\end{table*}

At first, we used Naïve sampling (Naïve) based on the 'gender' category since the WER values for this category appeared to differ significantly. We achieved data balance by randomly duplicating instances until we had an approximately equal number of male and female records. However, we found that naïve sampling did not improve the balance of the other categories. Therefore, we turned to the Synthetic Minority Over-sampling Technique (SMOTE) \cite{Smote} in the final stage. 

The SMOTE algorithm aims to tackle the issue of imbalanced data by creating synthetic observations for minority classes. It does not simply repeat the existing samples but rather creates similar examples that improve performance accuracy. It starts with choosing an instance in the minority class and computing the difference of feature vectors with neighboring observations. After that, the algorithm defines a region of \(k\) nearest neighbors around the selected instance. Next, the algorithm calculates the difference between observations and multiplies the difference vector by a random number from the range (0, 1), thus having a new synthesized sample. We do the resampling for every category one by one assuming the improvement in results. 

The statistics for evaluation data compiled using oversampling techniques along with initial samples are shown in Table \ref{tab:transposed_table_complete}. The average duration of each sample used for the evaluation of multilingual ASR systems corresponds to 2 minutes with the same textual content. Therefore, the robustness of ASR systems to long-form audio is an important consideration in the development and deployment of ASR technology. 

\subsubsection{Evaluation strategy}

For the evaluation of both models, we use the Word Error Rate (WER), a standard metric for ASR.  The Word Error Rate depicts the percentage of incorrectly recognized words and is calculated as follows:
\begin{equation}
    \small
    WER = \frac{S + D + I}{N},
\end{equation}
where \( S \) stands for number of substitutions, \( D \) for the number of deletions, \( I \) is the number of insertions, and \( N \) for the number of words in the reference transcription. In the current paper, we report the WER for comparison purposes with the literature, and we also report the Character Error Rate (CER) in \url{https://biasinai.github.io/asrbias/}.
 %since employing it enables us to operate uniformly and guarantees comparability with evidence from the mentioned publications. 
This allows us to compare results objectively and to identify performance biases in the 4 ASR systems.

\section{Results and Analysis}

In this section, we present a comprehensive analysis of Word Error Rate (WER) within distinct categories as provided by CCD V2. These categories include gender (Section 4.1), skin tone (Section 4.2), age groups (Section 4.3), and geo-location (Section 4.4). As previously mentioned, our experimentation involved the use of three Whisper ASR variants: Medium (769 million parameters), Large (1550 million parameters), and Large-v2 (1550 million parameters)\footnote{https://huggingface.co/openai/whisper-large-v2} (which maintains the same parameter count but benefits from extended training with regularization). Additionally, we utilized the MMS ASR system\footnote{https://huggingface.co/facebook/mms-1b-all} with 965 million parameters. 

\subsection{Gender analysis}

\begin{figure*}[!th]
    \centering
    \includegraphics[width=\textwidth]{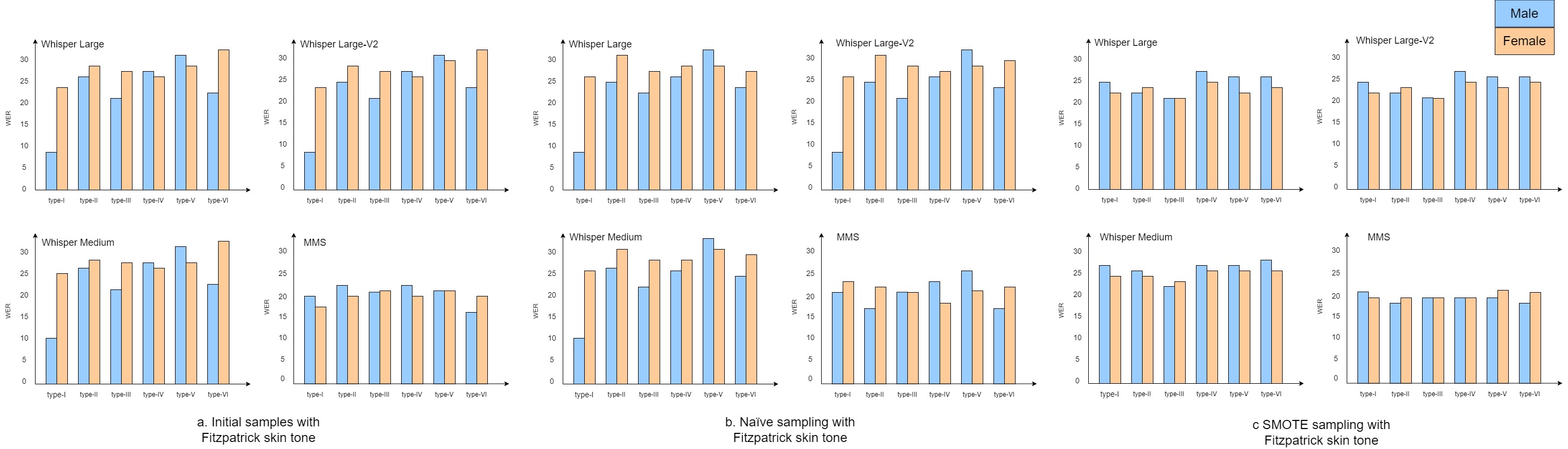}
    \caption{Bar plots depicting Whisper and ASR performance across the Fitzpatrick skin-tone scale, ranging from type-I to type-VI, for both male and female genders, with results for initial samples, naïve sampling, and SMOTE sampling}
    \label{fig:maingen2}
\end{figure*}

\begin{figure}[!t]
    \centering
    \includegraphics[width=0.5\textwidth]{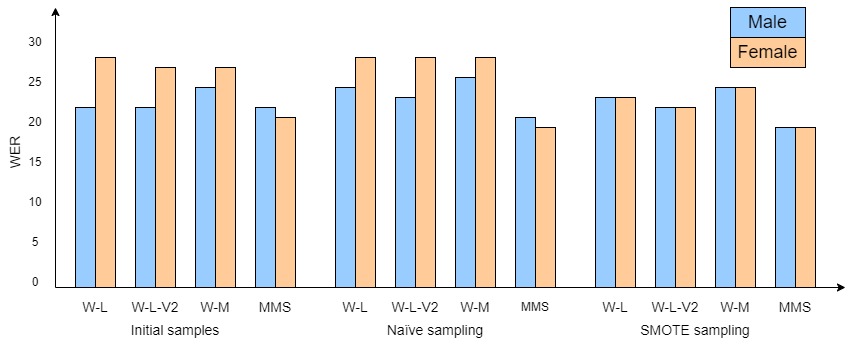}
    \caption{Bar-plots demonstrating performance of multilingual ASR systems using Whisper ASR variants and MMS for impact on male and female genders using WER under three sampling methods, initial, naïve and SMOTE. Whisper ASR variants are indicated as, Whisper-Large (W-L), Whisper-Large-V2 (W-L-V2), and Whisper-Medium (W-M).}
    \label{fig:main_gen}
    %\vspace{-1em}
\end{figure}

\begin{table}[t!]
\centering
\resizebox{0.4\textwidth}{!}{% Resize table to fit the text line width
\begin{tabular}{|l|l|l|l|l|}
\hline
Method        & W-L      & W-L-V2 & W-M  & MMS    \\ \hline
Initial     & 0.00022  & 0.00018 & 0.0011  & 0.195  \\ \hline
Naïve       & 2.07e-17 & 1.54e-17 & 1.45e-11 & 0.177  \\ \hline
SMOTE       & 0.676 & 0.603 & 0.778 & 0.563\\ \hline
%0.581    

%0.878    
 
%0.795

%0.786
\end{tabular}
}
\caption{$p$-values for Whisper ASR variants and MMS for the Gender category across Initial, Naïve and SMOTE datasets. Whisper ASR variants are indicated as, Whisper-Large (W-L), Whisper-Large-V2 (W-L-V2), and Whisper-Medium (W-M).}
%{\bf We could have done it for the combined genedred categories...}}
\label{table:asr_gender_by_method}
\vspace{-1em}
\end{table}

We illustrate the performance of ASR systems for the Portuguese language, on the gender subgroups 'Male' and 'Female'. From Figure \ref{fig:main_gen}, we observe a subtle gender bias when examining the Whisper ASR variants, which favors males in both the Initial and naïve sampling techniques. However, the use of SMOTE sampling results in a more balanced ASR performance between the gender subgroups. Notably, the MMS system outperforms the Whisper ASR variants, exhibiting comparatively balanced WER across both genders. 
As illustrated in Figure \ref{fig:main_gen}, we observe the absence of significant performance disparities between male and female genders. 

In addition to analyzing WERs, we also conducted a p-value analysis to assess the statistical significance of gender-related differences. In the examination of Table \ref{table:asr_gender_by_method}, we observed that the p-values for Whisper ASR variants applied to initial samples and Naïve sampling fell below the significance threshold of 0.05. This suggests that statistically significant differences exist between male and female gender categories in these cases. Conversely, the p-value statistics for the MMS approach consistently exceeded 0.05, indicating that there are no significant performance variations across both genders regardless of the sampling method. Regarding SMOTE sampling, the p-values for all ASR systems exceeded the 0.05 threshold, signifying evidence of mitigating gender biases in this context.

%{\bf !!!!Mention p-values? } We can do the $p$-value study on the combination with gender...

\begin{figure}[!th]
    \centering
    \includegraphics[width=0.5\textwidth]{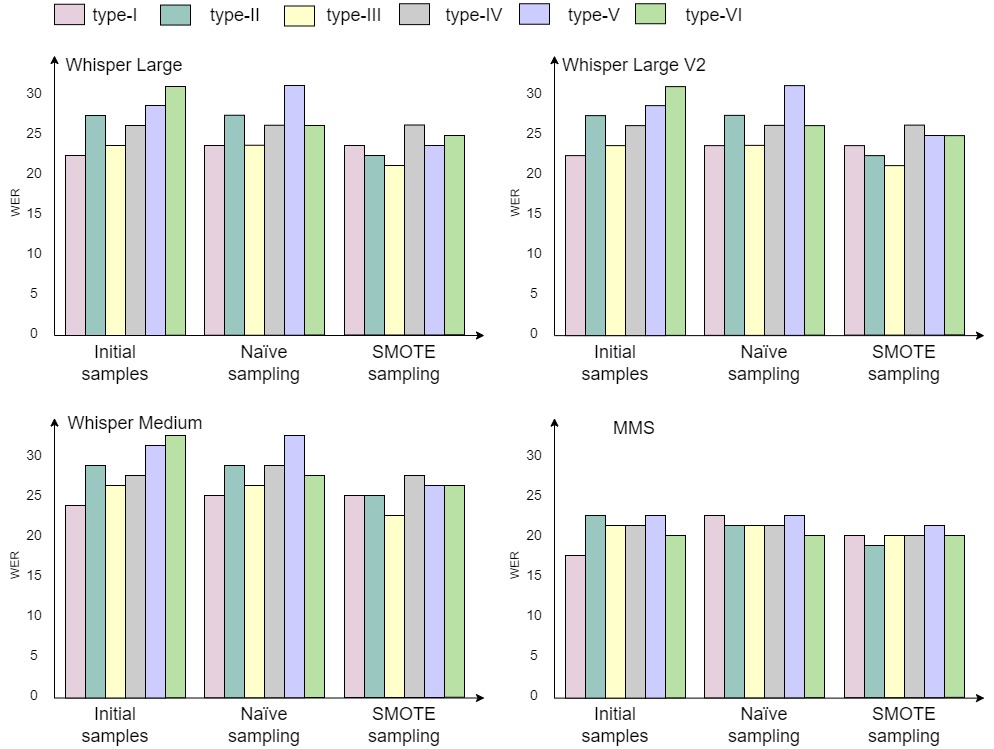}
    \caption{Bar-plots illustrating the distribution of mean WER for Fitzpatrick skin tone scales across Initial, naïve, and SMOTE sampling methods.}
    \label{fig:mainskin}
\end{figure}

\begin{figure*}[!h]
    \centering
    \includegraphics[width=\textwidth]{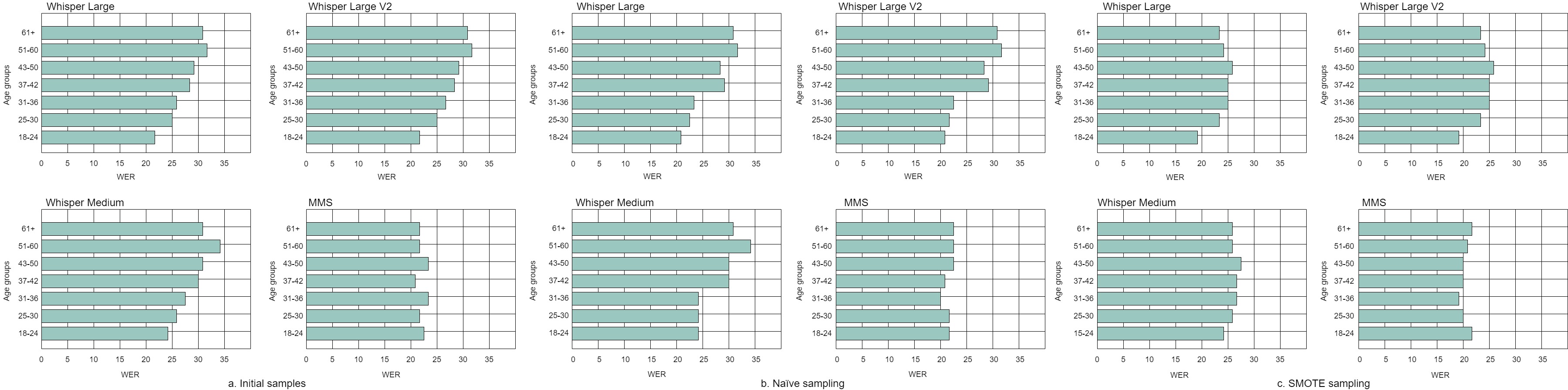}
    \caption{Bar-plots illustrating distribution of WER for age groups categorized into five sub-sets (18-24, 25-30, 31-36,  37-42, 42-50, 51-60, 61+) across initial, naïve and SMOTE sampling methods.}
    \label{fig:mainage}
\end{figure*}

\begin{figure*}[!h]
    \centering
    \includegraphics[width=\textwidth]{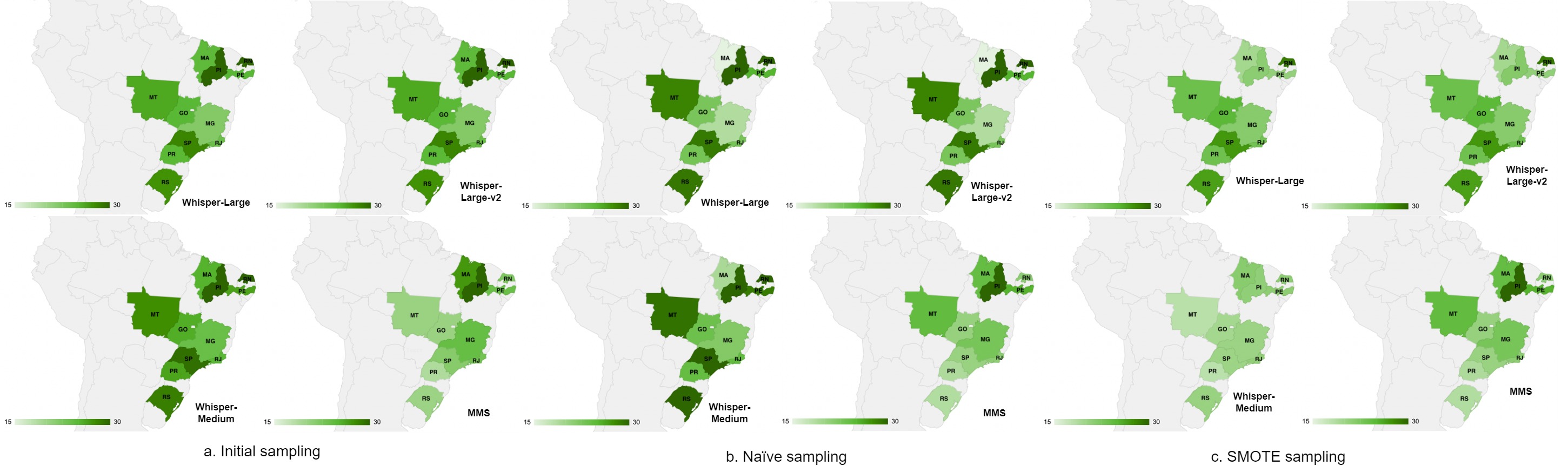}
    \caption{The visualization of mean WER distribution in each Portuguese state. The abbreviations of states are as follows: RN - Rio Grande do Norte, SP - Sao Paulo, RS - Rio Grande do Sul, GO -Goias, MT - Mato Grosso, PR - Parana, RJ - Rio de Janeiro, MG - Minas Gerais, PI - Piaui, PE - Pernambuco, MA - Maranhao.}
    \label{fig:mainlocation}
\end{figure*}

After this, we extended our study of ASR systems with the distribution of WER performances concerning skin tone as measured by the Fitzpatrick skin type and gender. This examination is depicted in Figure \ref{fig:maingen2}. Significant disparities are evident across different skin tone types between male and female individuals. Specifically, within the Whisper ASR variants, notable performance differences are observed for skin-tone type-I and type-VI. In these cases, the male subgroup exhibits better WER compared to the female subgroup, particularly in the context of initial samples and naïve sampling approaches. Moreover, the MMS ASR system demonstrates a relatively even distribution of WER across all skin-tone types and outperforms all variants of the Whisper ASR. It is worth highlighting that, across all the ASR systems under examination, the use of SMOTE sampling has consistently played a role in mitigating performance disparities, leading to more balanced outcomes across gender subgroups.

\subsection{Skin-tone analysis}

%\vspace{-1}

We also examine the impact of ASR performance within sub-categories using categorized by Fitzpatrick skin tone type, without conditioning on other meta-attributes. Figure \ref{fig:mainskin} shows the relative performance variations across various sampling techniques applied to ASR systems. Notably, we observe that individuals with skin types I to III demonstrate comparatively better WER than those with skin type IV. This observation sheds light on potential racial biases in ASR systems, where greater skin-type variations often indicate darker skin colors.

However, amidst these disparities, the MMS ASR system stands out with evenly distributed WER measures across all skin-type scales. When assessing the differences introduced by sampling approaches, initial samples, and naïve sampling reveal disparities among skin-tone subgroups. In contrast, the consistent use of SMOTE sampling proves effective in mitigating discrepancies across all the ASR systems under investigation.

\subsection{Age group analysis}

In Figure \ref{fig:mainage}, we present an age group analysis of the Portuguese language for ASR systems using three different sampling techniques: initial samples, naïve sampling, and SMOTE sampling. Across all the sampling methods, the MMS ASR system consistently maintains WER measures below 25\% for all age groups, exhibiting a relatively even distribution of WER values. In contrast, the Whisper ASR variants demonstrate disproportionate WER measures, particularly noticeable between the age groups of 18-36 and 36+. Moreover, the performance of the Whisper ASR degrades as age groups increase. However, the utilization of SMOTE sampling significantly improves the WER of the Whisper systems, bringing it to an overall 25\%. 

This distinctively highlights the positive impact of SMOTE sampling in reducing performance disparities across various age groups for both the Whisper and MMS ASR systems.

\subsection{Geo-location analysis}

Figure \ref{fig:mainlocation} provides a comprehensive examination of the impact of different sampling techniques on ASR performance disparities across various regions in Brazil. Notably, when considering the Whisper ASR system, regions such as São Paulo (SP), Piauí (PI), Rio Grande do Norte (RN), and Rio Grande do Sul (RS) are notably affected by performance differences, regardless of whether initial samples or naïve sampling methods are employed. These regions exhibit significant variations in WER compared to other regions. 
Overall, the MMS ASR system displays a more even distribution of evaluation measures across all sampling approaches and generally outperforms the Whisper ASR variants. Furthermore, it is notable to highlight that, despite observing proportionate WERs across most regions in Brazil, the MMS ASR system experiences a decline in performance specifically in the Piauí (PI) region for all sampling approaches. 

Even after the application of SMOTE sampling, the Whisper ASR variants continue to exhibit consistently higher WER values in the Rio Grande do Norte (RN) region. However, SMOTE sampling effectively mitigates WER discrepancies in the Piauí (PI) region. This underscores the distinct challenges posed by regional variations in ASR performance and underscores the potential of SMOTE sampling in addressing these disparities.

%a. Metrics used
%b. Graphical representations
%c. Analysis on each subgroups for both languages
%    I. Skin-tone
%    II. Gender
%    III. Location
%    IV. Age groups
%d. Port Vs Spanish
%e. MMS Vs Whisper 

\section{Discussion and limitations}  

Our results reveal that all 4 models show mild  WER performance disparities when considering the individual subgroups of the categories `Gender', `Age', `Skin Tone Color', and `Geo-location', with a consistently better performance of the MMS model over the three Whisper models. However, when analyzing the gendered subcategories of `Age', `Skin Tone Color', and `Geo-location', we observe significant differences in WER, with a noticeable bias that privileges the `Male' subgroup; see additional results in \url{https://biasinai.github.io/asrbias/}. 

Our study also shows that oversampling approaches can alleviate these disparities between the two gender subgroups. This is particularly evident in Figure~\ref{fig:main_gen}, where WER performances are balanced for the `Male' and `Female' subgroups over the 4 models considered. The same trend was also observed for the other gendered categories and with respect to the Character Error Rate (CER) in the link provided earlier. The study shows that performances of Whisper variants demonstrate higher sensitivity to the number of model parameters, whereas the MMS system, despite having 40\% fewer parameters than Whisper Large, exhibits better robustness over the various categories. 

Despite promising, these results naturally ask for similar comparisons with respect to other performance and bias metrics. Another limitation of our study is that it was carried out solely on the CCD V2. In \cite{artiebias}, the Artie Bias Corpus is curated as a subset of the Mozilla Common Voice corpus. It includes demographic tags for {age, gender, and accent}, which allows for the examination of disparities in the English language. It is imperative to construct bias-focused datasets using publically available resources for Portuguese. 

Furthermore, we can also extend this investigation to other state-of-the-art multilingual ASR systems such as Universal speech model \cite{ggleusm}, ASR2K \cite{asr2k}, and DeepSpeech \cite{deepspeech} and on other tasks ({\it e.g.,} speaker verification \cite{svbias}). Also, we only experimented with the original SMOTE \cite{Smote} framework, but improvements could be obtained with dedicated versions, {\it e.g.,} \cite{Smote2}, \cite{Smote3}, \cite{Smote4}. Our study focused on the Portuguese language but we are currently extending it to other languages. Finally, these results ask for a thorough analysis to detect the speech meta-features that trigger the disparate behavior of these ASR systems. For instance, correlational features among skin-tone scale and voice-timber in speech utterances affect the disparity gap in performance.

%-other models/tasks
%- Need for tools and evaluation metrics to quantify biases
%- Close set analysis Vs Causal conversation
%-limits of the study

%\section{future perspectives}

\section{Conclusion}

In this work, we presented an extensive study of recent ASR systems, namely, Whisper and MMS, in the light of stereotypical biases such as gender, age, skin tone, and geo-location, for the Portuguese language. Despite observing mild performance disparities concerning individual categories such as  `Age', `Skin Tone Color', and `Geo-location', we empirically show significant performance differences between the `Male' and `Female' subgroups. The first observation was to notice the imbalance in the various distributions, and that a naïve oversampling may further contribute to disparate performance behavior. This motivated us to employ SMOTE, and our results attested that oversampling technique has an overall beneficial impact in reducing performance differences. We also discuss some limitations of our study along with future work.

\bibliography{custom}
\bibliographystyle{acl_natbib}

\end{document}